\documentclass[10pt,twocolumn,letterpaper]{article}
\usepackage{wrapfig}
\usepackage{cvpr}              

\usepackage{graphicx}
\usepackage{amsmath}
\usepackage{amssymb}
\usepackage{booktabs}
\usepackage{xspace}
\usepackage{bm}
\usepackage{amsmath}
\usepackage{floatrow}
\newfloatcommand{capbtabbox}{table}[][\FBwidth]

\DeclareMathOperator{\Tr}{Tr}
\newcommand{\NCD}{Novel Class Discovery\xspace}
\newcommand{\SL}{Spacing Loss\xspace}
\DeclareMathOperator*{\argmin}{arg\,min}
\usepackage{enumitem}

\usepackage{algorithm}
\usepackage[noend]{algpseudocode}

\algdef{SE}[DOWHILE]{Do}{doWhile}{\algorithmicdo}[1]{\algorithmicwhile\ #1}%

\usepackage[pagebackref,breaklinks,colorlinks]{hyperref}
\usepackage[dvipsnames,table,xcdraw]{xcolor}

\usepackage[capitalize]{cleveref}
\Crefname{equation}{Eq.}{Eqs.}
\Crefname{figure}{Fig.}{Figs.}
\Crefname{tabular}{Tab.}{Tabs.}
\Crefname{section}{Sec.}{Secs.}
\Crefname{table}{Tab.}{Tabs.}

\def\cvprPaperID{17} 
\def\confName{CVPR}
\def\confYear{2022}
\definecolor{Gray}{gray}{0.95}

\begin{document}

\title{\vspace{-20pt}Spacing Loss for Discovering Novel Categories \vspace{-15pt}}

\author{K J Joseph$^{1,2}$~~~~ Sujoy Paul$^{2}$~~~~ Gaurav Aggarwal$^{2}$~~~~ Soma Biswas$^{3}$~~~~ Piyush Rai$^{2,4}$~~~~ \\ Kai Han$^{2,5}$~~~~ Vineeth N Balasubramanian$^{1}$\\
$^1$Indian Institute of Technology Hyderabad \hspace{1em} $^2$Google Research \hspace{1em} $^3$Indian Institute of Science \\ $^4$Indian Institute of Technology Kanpur \hspace{1em} $^5$The University of Hong Kong\\
{\tt\small \{cs17m18p100001,~vineethnb\}@iith.ac.in, somabiswas@iisc.ac.in,} \\
{\tt\small \{sujoyp,~gauravaggarwal,~piyushrai,~kaihanx\}@google.com}
\vspace{-10pt}
}
\maketitle

\begin{abstract}
Novel Class Discovery (NCD) is a learning paradigm, where a machine learning model is tasked to semantically group instances from unlabeled data, by utilizing labeled instances from a disjoint set of classes. 
In this work, we first characterize existing NCD approaches into single-stage and two-stage methods based on whether they require access to labeled and unlabeled data together while discovering new classes.
Next, we devise a simple yet powerful loss function that enforces separability in the latent space using cues from multi-dimensional scaling, which we refer to as Spacing Loss. 
Our proposed formulation can either operate as a standalone method or can be plugged into existing methods to enhance them. We validate the efficacy of Spacing Loss with thorough experimental evaluation across multiple settings on CIFAR-10 and CIFAR-100 datasets. 
\end{abstract}

\section{Introduction}
Availability of large amount of annotated data has fueled unprecedented success of deep learning in various machine learning tasks \cite{bulat2020toward,duan2019centernet,sauer2021projected,joseph2021towards,mohan2021efficientps,tolstikhin2021mlp}. Though human learners also require various levels of supervision throughout their lifetime, we make use of the bulk of knowledge acquired so far to make intelligent choices, which guides effective learning. 
Drawing a parallel to the machine learning problem of image classification, it is natural to expect a model trained on a huge number of labeled classes (e.g., 1000 classes in ImageNet dataset \cite{russakovsky2015imagenet}) to give meaningful representations to identify and differentiate instances of novel categories. This is the basis for the research efforts in Novel Class Discovery (NCD) setting \cite{Hsu18_L2C,Hsu19_MCL,han2019learning,han2019automatically,zhong2021neighborhood,Fini2021unified,zhao2021novel}. Given access to labeled training data from a set of classes, an NCD model identifies novel categories from an unlabeled pool containing instances from a disjoint set of classes. 

As the nascent field of \NCD continues to evolve, we introduce a categorization of existing NCD methods based on the data that is required to train them. 
\textit{Single-stage} NCD models can access labeled data and unlabeled data together while discovering novel categories from the latter. \textit{Two-stage} NCD models can access labeled and unlabeled data only in stages. 
Each of these settings has a wide practical applicability. 
Consider a marine biologist who studies about various kinds of organisms in the ocean, from images captured by under-water vehicles \cite{katija2021fathomnet,katija2021visual}. While analysing these images for novel categories in their lab, it would be ideal to make use of any annotated data that they might have already collected overtime. Hence, a single-stage NCD methods would be ideal for their setting. 
Contrastingly, consider an autonomous robot that can assist the visually impaired \cite{kulyukin2005robocart,kulyukin2006robot}. While being operational, it would be great for the robot to discover and identify instances of novel categories in the environment, so that it can alert its users. In this scenario, it is not practical to reuse all labeled instances that the robot was trained on in its factory, while discovering novel categories. A two-stage NCD method is more desired in this setting.

A common theme in most NCD methodologies is to learn a feature extractor using the labeled data and use clustering \cite{Hsu18_L2C,Hsu19_MCL,han2019learning}, psuedo-labelling based learning \cite{han2019automatically,Fini2021unified} or contrastive learning \cite{zhong2021neighborhood,jia21joint} to identify classes in the unlabeled pool. 
In contrast, we propose a novel \textit{Spacing Loss} which ensures separability in the latent space of feature extractor, for the labeled and unlabeled classes.
This is achieved by transporting semantically dissimilar instances to equidistant areas in the latent space, identified via multi-dimensional scaling \cite{webb1995multidimensional}.
We note that our proposed loss formulation is orthogonal to the existing methodologies, and can easily complement these methods. Our experimental evaluation on CIFAR-10 \cite{krizhevsky2009cifar} and CIFAR-100 \cite{krizhevsky2009cifar} datasets suggests that the models trained with the proposed \SL achieve state-of-the-art performance when compared to two-stage NCD methods. Further, when combined with single-stage methodologies, 
our loss formulation improves each of them consistently. 

The standard strategy to evaluate NCD methods is to train the model on a subset of classes from a classification dataset and evaluate its performance on the remaining classes. Complementing existing protocols, we introduce a new split where the number of classes in the labeled pool is significantly lower than the number of classes in the unlabeled pool. Such a protocol aligns more closely with the real-world scenarios, where the number of classes in the labeled and unlabeled pool might be heavily imbalanced. 

\vspace{3pt}
\noindent To summarize, the key contributions of our work are:
\begin{itemize}[leftmargin=*,topsep=0pt, noitemsep]
\item We propose Spacing Loss, which enforces separability in the latent space, for the challenging problem of novel category discovery. 
\item 
We evaluate our proposed approach on benchmark datasets for novel category discovery, under both single- and two-stage settings, consistently outperforming existing methods.
\end{itemize}

\section{Novel Class Discovery Methods}

\noindent \textbf{Two-stage Methods } \
Early methods in \NCD \cite{Hsu18_L2C,Hsu19_MCL,han2019learning} operate in a phased setting. In the first phase, the model learns from the labeled data, and in the subsequent phase, it discover novel categories from the unlabeled pool. MCL \cite{Hsu19_MCL} and KCL \cite{Hsu18_L2C} learn a binary similarity function using meta-learning in the first phase, and use this in the category discovery phase. DTC \cite{han2019learning} first learns a feature extractor on the labeled data. In the next stage, these features are used to initialise a clustering algorithm \cite{xie2016unsupervised}, which further fine-tunes these representations using the unlabeled data, thereby improving class discovery.

\noindent \textbf{Single-stage Methods } \
More recent efforts in NCD \cite{han2019automatically,zhong2021neighborhood,Fini2021unified} use labeled and the unlabeled data together to discover novel categories. RS \cite{han2019automatically,han21autonovel}, NCL \cite{zhong2021neighborhood} and OpenMix \cite{zhong2021openmix} first use RotNet \cite{komodakis2018unsupervised} to self-supervise on the labeled and unlabeled data. Then, RS \cite{han2019automatically} uses pseudo-labels from ranking-statistics method to learn an unlabeled head.  NCL \cite{zhong2021neighborhood} and Jia \etal \cite{jia21joint} find that contrastive learning improves class discovery and OpenMix \cite{zhong2021openmix} uses mix-up \cite{zhang2018mixup} to generate more training data to guide class discovery. UNO \cite{Fini2021unified} finds that a unified loss function enhances the synergy between the learnings from labeled and the unlabeled data. Zhao and Han \cite{zhao2021novel} proposes to focus on fine-grained local cues in images to enhance discrimination\footnote{As NCD is a nascent field, we will maintain an updated list of methods here: \href{https://github.com/JosephKJ/Awesome-Novel-Class-Discovery}{https://github.com/JosephKJ/Awesome-Novel-Class-Discovery}.}.



\vspace{10pt}
\section{Spacing Loss}
\begin{figure}
\centering
\includegraphics[width=1\columnwidth]{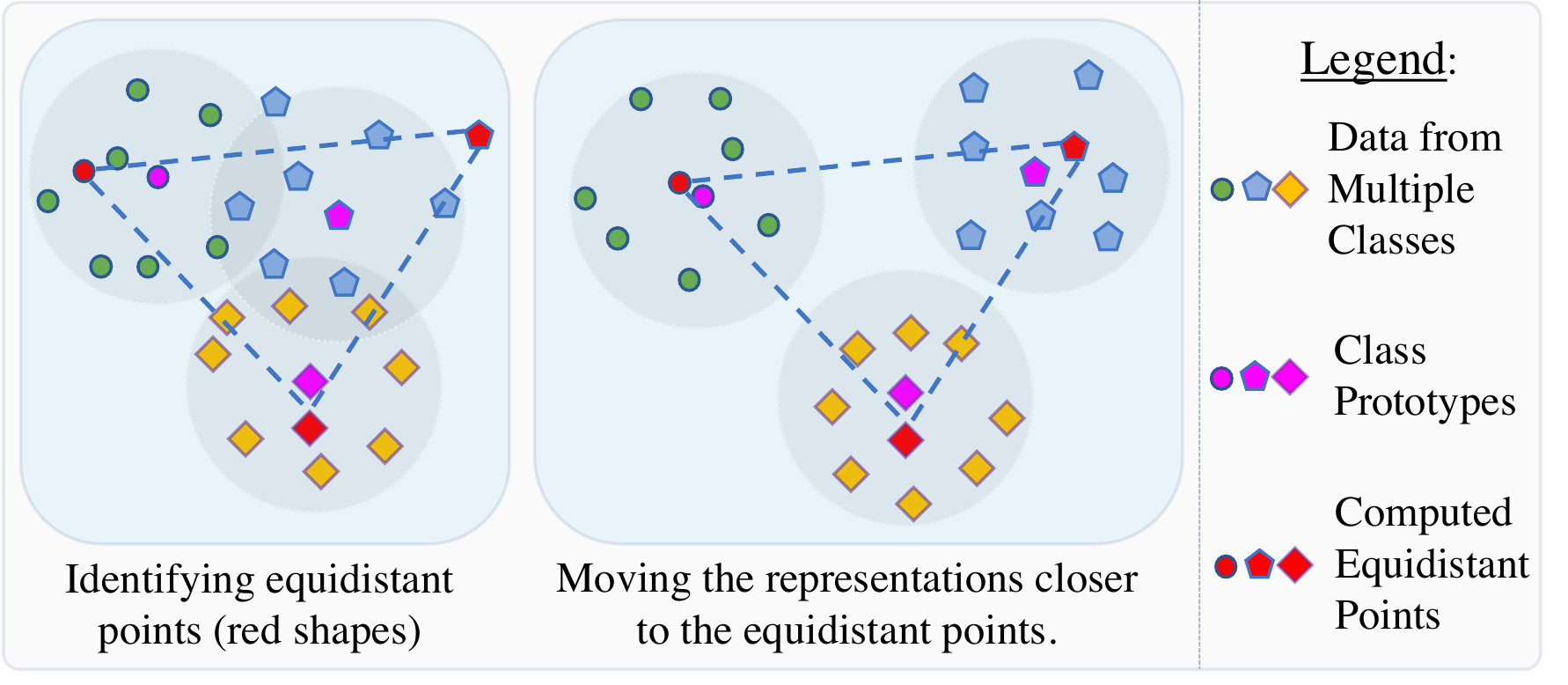}
\vspace{-20pt}
\caption{The figure illustrates how latent space is adapted by the proposed \SL. 
Latent representations from different classes are shown in different shapes. As the model is bootstrapped with labeled data, the latent representations from the unlabeled data will have reasonable semantic grouping. We further enhance the separability in the latent space by identifying equidistant points (shown in red) and then moving the latent representations to these identified locations, effectively ensuring spacing between the classes of interest. 
\vspace{-15pt}
}
\label{fig:illustration}
\end{figure}

Learning to adapt the latent representations of a model, such that semantically identical samples would share nearby locations in the latent manifold, while semantically dissimilar samples are spaced apart, would be ideal for discovering novel classes. 
Such a subspace shaping should evolve as latent representations mature. 
Two characteristics would be ideal in such a setting: 1) the ability to transport similar samples to locations equidistant from other dissimilar samples in the latent manifold, 2) the datapoints having the ability to refresh their associativity to a group as the learning progresses. 
We propose a simple yet effective methodology that accommodates both aspects.
\Cref{fig:illustration} illustrates how the latent space is adapted using the proposed \SL.  
While learning to discover classes, we identify locations in latent space  (in red), which are equidistant from each other. Next, we enforce the latent representations from unlabeled data to be transported to nearest of such points. Each latent representation can change their membership to a specific group as the learning progresses. This flexibility along with the weak regularization enables us to learn a well-separated latent representation. A simple non-parametric inference in this space can help us to discover categories. 
We summarize how equidistant locations is identified in \Cref{sec:MDS}, 
followed by how latent space is adapted in \Cref{sec:adaptive_spacing},
concluding with the overall objective in \Cref{sec:overall_objective}.


\subsection{Finding Equidistant Points in the Latent Space} \label{sec:MDS}
Let us consider a feature extractor $\Phi:\mathbb{R}^{w \times h \times 3}\rightarrow \mathbb{R}^{z}$, which takes an input image and generates a $z$ dimensional latent representation. 
We identify $c$ prototypes,
$\bm{P}= \{\bm{p}_1,\cdots,\bm{p}_c\}$, from these latent representations, where $c$ is the total number of classes under consideration.
These prototypes can be initialised using a simple centroid based strategy. We identify equidistant points, $\bm{P}^e= \{\bm{p}_1^e \cdots \bm{p}_c^e\}$, in this latent space which are guaranteed to be far apart at-least by the largest pair-wise distance between these prototypes. These equidistant points serve as anchors to which the corresponding centroids and its associated data would be progressively shifted to while the learning progresses.

Let $p_{dist}$ be the largest pair-wise distance between the prototypes. We first construct a $c \times c$ \textit{dissimilarity matrix} $\bm{\Delta}$ as follows: all entries but for the diagonals are set to $\delta_{ij} = \alpha \times p_{dist}$, where $\alpha > 1$. The diagonal elements $\delta_{ij}$, of $\bm{\Delta}$ are set to $0$. Hence, $\bm{\Delta}$ is symmetric, non-negative and hollow by construction. Given  a dataset $\bm{D}$, we seek to find $\bm{P}^e= \{\bm{p}_1^e \cdots \bm{p}_c^e\}$ where each $\bm{p}_i^e \in \mathbb{R}^z$, such that distance between any $\bm{p}_i^e$ and $\bm{p}_j^e$ is approximately $\delta$: ie. $d_{ij}(\bm{P}^e) \approx \delta_{ij}$. $d_{ij}(\bm{P}^e)$ corresponds to the distance between $\bm{p}_i^e$ and $\bm{p}_j^e$ in euclidean space.
We can formulate the objective to learn $\bm{P}^e$ as follows:
\begin{equation}
    \sigma(\bm{P}^e) = \sum_{i < j \leq c} w_{ij} (d_{ij}(\bm{P}^e) - \delta_{ij})^2,
    \label{eqn:objective}
\end{equation}
where $\bm{W}$ is a symmetric, non-negative and hollow matrix of weights $w_{ij}$, which captures the relative importance. For simplicity, we weigh each $\bm{P}^e_i$ equally. 
\noindent As finding an analytical solution to minimize \Cref{eqn:objective} is intractable, an iterative majorization algorithm \cite{webb1995multidimensional,borg2005modern,de2005applications} is used. We seek to find a manageable surrogate function $\tau(\bm{P}^e, \bm{Y})$, which majorizes $\sigma(\bm{P}^e)$, i.e., $\tau(\bm{P}^e, \bm{Y}) > \sigma(\bm{P}^e)$, with the initial supporting points $\bm{Y}$.
We can rewrite \Cref{eqn:objective} as follows:
\begin{equation}
    \sigma(\bm{P}^e) = \sum_{i < j} d_{ij}^2(\bm{P}^e) + \sum_{i < j} \delta_{ij}^2 - 2\sum_{i < j}\delta_{ij}d_{ij}(\bm{P}^e).
    \label{eqn:objective_rewrite}
\end{equation}

\noindent The first term is a quadratic in $\bm{P}^e$ and can be expressed as $\Tr {\bm{P}^e}^T \bm{V} \bm{P}^e$, where $\bm{V}$ has $v_{ij} = -w_{ij}$ and $v_{ii} = \sum w_{ij}$  \cite{de2005applications}.
The second term is a constant, say $k$, and the third term can be bounded as follows:

\vspace{-10pt}
\begin{align}
\sum_{i < j}\delta_{ij}d_{ij}(\bm{P}^e) &= \Tr {\bm{P}^e}^T \bm{B}(\bm{P}^e) \bm{P}^e \nonumber \\
& \geq  \Tr {\bm{P}^e}^T \bm{B}(\bm{Y}) \bm{Y},
\label{eqn:objective_matrix_notation}
\end{align}

\noindent where $\bm{B}(\bm{Y})$ has 

\vspace{-10pt}
\begin{align}
b_{ij} &= \begin{cases}
    \frac{\delta_{ij}}{d_{ij}(\bm{Y})}, & \text{for } d_{ij}(\bm{Y}) \neq 0, i \neq j \\
    0, & \text{for } d_{ij}(\bm{Y}) = 0, i \neq j 
  \end{cases} \text{~and }  \nonumber \\
b_{ii} & =  -\sum_{j=1, j \neq i}^c b_{ij}.
  \label{eqn:b}
\end{align}


\noindent The proof of this inequality follows \cite{de2005applications,borg2005modern}. Hence, the surrogate function that majorizes $\sigma(\bm{P}^e)$ is as follows: 
\begin{equation}
    \tau(\bm{P}^e, \bm{Y}) = \Tr {\bm{P}^e}^T \bm{V} \bm{P}^e + k - 2~\Tr {\bm{P}^e}^T \bm{B}(\bm{Y}) \bm{Y}.
    \label{eqn:surrogate_objective}
\end{equation}
\begin{algorithm}[H]
\small
\caption{\textsc{GetEquidistantPoints}}
\label{algo:mds}
\begin{algorithmic}[1]
\Require{Prototype vectors: $\bm{P}= \{\bm{p}_0 \cdots \bm{p}_c\}$, Small constant $\epsilon$.
}
\Ensure{Equidistant points: $\bm{P}^e$
}
\State $p_{dist} \leftarrow$ maximum distance between all prototypes in $\bm{P}$.
\State Compute $\bm{\Delta}$ from $p_{dist}$.

\State Initialize $\bm{P}^e$ randomly.

\Do
    \State $\bm{Y} \leftarrow \bm{P}^e$
    \State $\bm{P}^e \leftarrow \argmin_{\bm{P}^e} \tau(\bm{P}^e, \bm{Y})$ \Comment{Defined in \Cref{eqn:surrogate_objective}}
\doWhile{$(\bm{Y} - \bm{P}^e) > \epsilon$}

\State \Return $\bm{P}^e$

\end{algorithmic}
\end{algorithm}
\vspace{-10pt}
\Cref{algo:mds} summarizes how $\bm{P}^e$ are computed by optimizing \cref{eqn:surrogate_objective}. In Line 2, we compute the dissimilarity matrix $\bm{\Delta}$ by using the maximum distance between the prototype vectors $\bm{P}$. First $\bm{P}^e$ is randomly initialised. Until there is negligible  change $\epsilon$ in $\bm{P}^e$, we update $\bm{P}^e$ to optimize the surrogate function $\tau(\bm{P}^e, \bm{Y})$. The resulting vectors in $\bm{P}^e$ are guaranteed to be equidistant from each other \cite{borg2005modern}. 

\subsection{Learning Separable Latent Space} \label{sec:adaptive_spacing}
Once the equidistant locations in the latent space $\bm{P}^e$ are identified, they can be used to enforce separation in the latent representations of images from different classes. As each latent representation matures with training, it might need to change its associativity with its initial group. We propose a novel formulation in \Cref{algo:LearningWithSpacing} that would allow for this flexibility during learning. The training essentially alternates between learning with pseudo-labels derived from class prototypes (Lines 6 - 8) and modifying the class prototypes themselves (Lines 11 - 15). In Line 1, we initialize the class prototypes $\bm{P}$ as the centroids of latents from $\Phi_{\bm{\theta}}$ using $k$-means \cite{macqueen1967some}. Based on the closeness to these prototypes, the class associativity of each image in a mini-batch is determined in Line 7. The feature extractor is updated to make the latent representations closer to these prototypes in Line 8. Using these newer features, the assignment is recomputed and the prototypes themselves are updated in Line 15. For each data-point $z_i$, its corresponding prototype $\bm{p}_{c_{\bm{z}_i}}$ is moved closer to the equidistant point $\bm{p}^e_{c_{\bm{z}_i}}$ and its current representation, controlled by a momentum parameter $\eta$. The parameter $\eta$ dampens with more instances of the specific class seen during training.

\begin{algorithm}
\small
\caption{\textsc{LearningWithSpacing}}
\label{algo:LearningWithSpacing}
\begin{algorithmic}[1]
\Require{Feature extractor: $\Phi_{\bm{\theta}}$, Data: $\bm{D} = \{\bm{X}_i\}$, \# of epochs: $e$.
}
\State Initialize class prototypes $\bm{P}= \{\bm{p}_0 \cdots \bm{p}_c\}$.
\State Identify equidistant points $\bm{P}^e= \{\bm{p}_0^e \cdots \bm{p}_c^e\}$ using Algo. \ref{algo:mds}.
\State Initialize assignment frequency $\bm{v} \leftarrow \bm{0}$; $|\bm{v}| = c$.
\For{each epoch $e$}
    \For{each minibatch ${\bm{X}} \subset \bm{D}$} 
        \State $\bm{Z} \leftarrow \Phi_{\bm{\theta}}(\bm{X})$ 
        \State $\bm{A} \leftarrow $ assign the nearest prototype from $\bm{P}$ for each $\bm{Z}$.
        \State Update $\bm{\theta}$ with MeanSquaredError($\bm{Z}$, $\bm{A}$).
        \State $\bm{Z} \leftarrow \Phi_{\bm{\theta}}(\bm{X})$ \Comment{Recompute $\bm{Z}$ with updated $\bm{\theta}$}
        \State $\bm{A}\leftarrow$ recompute prototype assign. for each new $\bm{Z}$.
        \For{$\bm{z}_i$ in $\bm{Z}$}
        \State $c_{\bm{z}_i} \leftarrow$ retrieve assignment index of $\bm{z}_i$ from $\bm{A}$.
        \State $\bm{v}[c_{\bm{z}_i}]\leftarrow \bm{v}[c_{\bm{z}_i}] + 1$
        \State $\eta \leftarrow \frac{1}{\bm{v}[c_{\bm{z}_i}]}$
        \State $\bm{p}_{c_{\bm{z}_i}} \leftarrow (1 - \eta) \bm{p}_{c_{\bm{z}_i}} + \eta (\bm{z}_i + \bm{p}^e_{c_{\bm{z}_i}})$ 
        \EndFor
    \EndFor
\EndFor
\end{algorithmic}
\end{algorithm}

\begin{figure*}
\begin{floatrow}
\capbtabbox{%
\resizebox{0.7\textwidth}{!}{%
\begin{tabular}{>{\kern-\tabcolsep}l|cccc|cccc<{\kern-\tabcolsep}}
\toprule
\rowcolor{Gray}\multicolumn{1}{c|}{Setting$\rightarrow$} & \multicolumn{4}{c|}{Imbalanced Class Split} & \multicolumn{4}{c}{Balanced Class Split} \\ \midrule
\rowcolor{Gray}\multicolumn{1}{c|}{Dataset Splits$\rightarrow$} & \multicolumn{2}{c|}{CIFAR-100-80-20} & \multicolumn{2}{c|}{CIFAR-100-20-80} & \multicolumn{2}{c|}{~~CIFAR-10-5-5~~} & \multicolumn{2}{c}{CIFAR-100-50-50} \\ \midrule
\rowcolor{Gray}\multicolumn{1}{c|}{Method} & CA & \multicolumn{1}{c|}{NMI} & CA & NMI & CA & \multicolumn{1}{c|}{NMI} & CA & NMI \\ \midrule
\multicolumn{1}{l|}{RS\cite{han2019automatically}} & 69.39 & \multicolumn{1}{c|}{0.6934} & 16.63 & 0.4493 & 89.72 & \multicolumn{1}{c|}{0.7724} & 47.72 & 0.5666 \\
\multicolumn{1}{l|}{RS + Spacing loss} & \textbf{73.16} & \multicolumn{1}{c|}{\textbf{0.7252}} & \textbf{26.37} & \textbf{0.4562} & \textbf{89.90} & \multicolumn{1}{c|}{\textbf{0.7764}} & \textbf{48.20} & \textbf{0.5712} \\ \hline
\multicolumn{1}{l|}{NCL \cite{zhong2021neighborhood}} & 81.01 & \multicolumn{1}{c|}{0.7883} & 19.82 & 0.4570 & 92.70 & \multicolumn{1}{c|}{0.8233} & 56.71 & 0.6355 \\
\multicolumn{1}{l|}{NCL + Spacing loss} & \textbf{85.11} & \multicolumn{1}{c|}{\textbf{0.7896}} & \textbf{35.60} & \textbf{0.5064} & \textbf{93.32} & \multicolumn{1}{c|}{\textbf{0.8364}} & \textbf{57.36} & \textbf{0.6432} \\ \bottomrule
\end{tabular}%
}
\label{tab:singelStage}
}{%
  \caption{We study the class discovery performance of single-stage NCD models across multiple settings in this table. Our proposed loss formulation can act as an add-on to existing methods, effectively enhancing their class discovery capability, even for severely skewed class distributions. 
  }%
}
\hspace{-70pt}
\ffigbox{%
  \frame{\includegraphics[width=0.49\columnwidth]{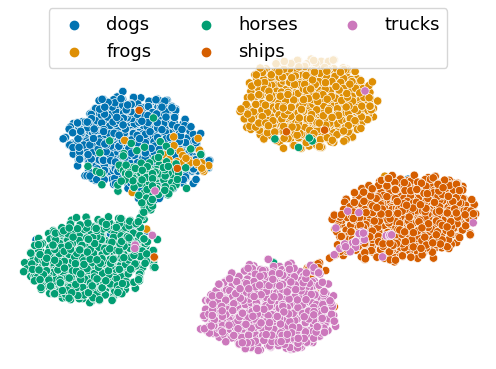}}
}{%
\captionsetup{width=.54\linewidth}
  \caption{Latent space of novel categories from CIFAR-10-5-5, trained using NCL + \SL. 
  }%
  \label{fig:tsne}
}
\end{floatrow}
\vspace{-15pt}
\end{figure*}

\vspace{-10pt}
\subsection{Overall Objective} \label{sec:overall_objective}
\vspace{-5pt}
So far, we have explained how the feature extractor $\Phi_{\bm{\theta}}$ is adapted by \SL. Our complete model extends this backbone with one head for the labeled data $F_{Lab} = \Phi_{Lab} \circ \Phi_{\bm{\theta}}$ and another for the unlabeled data $F_{Ulab} =\Phi_{Ulab} \circ \Phi_{\bm{\theta}}$. $F_{Lab}$ is learned with the labeled examples.  $F_{Ulab}$ is learned with pairwise pseudo labels derived from cosine-similarity \cite{zhong2021neighborhood} between its latent representations. We also enforce consistency in prediction with an augmented view of each image \cite{zhong2021neighborhood,han2019automatically,han2019learning,zhao2021novel} to enhance learning. 
While learning a two-stage model, we first learn $F_{Lab}$ using cross entropy loss with labeled data and then learn $F_{Ulab}$ with these auxiliary losses and \SL operating in the latent space. Labeled and unlabeled data, along with all the losses are used to learn the single-stage model. During inference, we do a $k$-means \cite{macqueen1967some} on the latent representations from the backbone network, to discover novel categories. 

\section{Experiments and Results}
Following existing NCD methods \cite{Hsu18_L2C,Hsu19_MCL,han2019learning,han2019automatically,zhong2021neighborhood,Fini2021unified,zhao2021novel}, we define splits on CIFAR-10 and CIFAR-100 to evaluate the efficacy of our method.
Clustering Accuracy \cite{han2019automatically} and NMI \cite{vinh2010information} are used as the evaluation criteria. 
We use ResNet-18 \cite{he2016identity} backbone and closely follow the hyper-parameter settings from Zhong \etal \cite{zhong2021neighborhood}.


\subsection{Two-stage Results}

\begin{wraptable}[12]{r}{0.30\textwidth}
\centering
\vspace{-10pt}
\resizebox{1\textwidth}{!}{%
\begin{tabular}{>{\kern-\tabcolsep}l|cc|cc<{\kern-\tabcolsep}}
\toprule
\rowcolor{Gray}\multicolumn{1}{c|}{{Datasets $\rightarrow$}} & \multicolumn{2}{c|}{{CIFAR-10}} & \multicolumn{2}{c}{{CIFAR-100}} \\ \midrule
\rowcolor{Gray}\multicolumn{1}{l|}{{Method}} & {CA} & {NMI} & {CA} & {NMI} \\ \midrule
\multicolumn{1}{l|}{K-means \cite{macqueen1967some}} & 65.5 & 0.422 & 66.2 & 0.555 \\
\multicolumn{1}{l|}{KCL\cite{Hsu18_L2C}} & 66.5 & 0.438 & 27.4 & 0.151 \\
\multicolumn{1}{l|}{MCL\cite{Hsu19_MCL}} & 64.2 & 0.398 & 32.7 & 0.202 \\
\multicolumn{1}{l|}{DTC\cite{han2019learning}} & 87.5 & 0.735 & 72.8 & 0.634 \\
\multicolumn{1}{l|}{RS*\cite{han2019automatically}} & 84.6 & 0.658 & 69.5 & 0.581 \\
\multicolumn{1}{l|}{NCL*\cite{zhong2021neighborhood}} & 60.5 & 0.479 & 59.5 & 0.428 \\
\multicolumn{1}{l|}{Spacing Loss} & \textbf{90.5} & \textbf{0.787} & \textbf{80.62} & \textbf{0.719} \\ \bottomrule
\end{tabular}%
}
\vspace{-7pt}
\caption{Regularization induced by \SL has better class discovery ability compared to baseline two-stage methods.
}
\label{tab:two_stage}
\end{wraptable}

In the first phase, we train the model on the labeled data from the first $5$ and $80$ classes from CIFAR-10 and CIFAR-100 datasets respectively for $200$ epochs. In the next phase, classes are identified from the unlabeled data guided by the \SL. \Cref{tab:two_stage} showcases the results. Our method consistently outperforms existing methods by a large margin, showing the efficacy of the proposed \SL. RS \cite{han2019automatically} and NCL \cite{zhong2021neighborhood} are adapted to the two-stage setting  for fair comparison (denoted by $^*$).  


\vspace{-5pt}
\subsection{Single-stage Results}
\vspace{-5pt}
A key characteristic of our proposed \SL is that the latent space regularization that it offers can effectively act as an add-on to existing methodologies. We showcase this capability while evaluating in single-stage setting. In Tab.~1, we organise different dataset splits based on the balance between the number of classes in labeled and unlabeled pool. The concise notation in Row 2 can be expanded as: dataset$-$total\_class\_count$-$labeled\_classes$-$
unlabeled\_classes. The latent space separation induced by \SL helps to improve the class discovery capability on all settings. It is interesting to note that the improvement is more pronounced in the more pragmatic setting, where the split of classes between the labeled and unlabeled pool is skewed. t-SNE \cite{van2008visualizing} visualization of backbone features in \cref{fig:tsne} shows good separation in these latent representations of novel categories in CIFAR-10-5-5 setting.

\vspace{-5pt}
\section{Enhancing Continual Learning with NCD}
\vspace{-5pt}
Continual learning setting aims to learn a single model which can incrementally accumulate knowledge across multiple tasks, without forgetting. Main-stream efforts in Continual Leaning \cite{li2017learning,rebuffi2017icarl,castro2018end,wu2019large,douillard2020podnet,liu2020mnemonics,rebuffi2017icarl,liu2020mnemonics,belouadah2019il2m,kj2020meta,rusu2016progressive,rajasegaran2019adaptive,rajasegaran2019random,abati2020conditional,liu2021adaptive} assume that the data which is introduced in each incremental task is fully annotated. Efforts in \NCD can help to relax this requirement, where the model could be tasked to identify classes from the instances of a new task automatically, based on the learnings that it already had. Then, these identified novel categories may be incrementally learned. 
We hope that the unification of these two streams of research would lead to a more pragmatic problem setting by building on their complementary characteristics. 


\vspace{-5pt}
\section{Conclusion}
\vspace{-5pt}
We characterise research efforts in the nascent \NCD setting into single-stage and two-stage methods, based on their data requirement during training. We further propose a simple yet effective method which enhances both these settings by enforcing separability in the latent representations. Our experimental analysis on multiple settings on two benchmark datasets corroborates with our assertions. Advancements in NCD can help continual learning models to operate in an open-world \cite{bendale2015towards,joseph2021towards}, where it can automatically identify novel categories and then incrementally learn them. We hope this pragmatic setting would be extensively explored in follow-up works.


\clearpage
{\small
\bibliographystyle{ieee_fullname}
\bibliography{egbib}
}

\end{document}